\documentclass[conference]{IEEEtran}
\IEEEoverridecommandlockouts
\usepackage{cite}
\usepackage{amsmath,amssymb,amsfonts}
\usepackage{algorithmic}
\usepackage{graphicx}
\usepackage{textcomp}
\usepackage{booktabs}
\usepackage{subcaption}
\usepackage[hidelinks]{hyperref}
\usepackage{multirow}
\usepackage{mathtools}
\usepackage{xcolor}
\def\BibTeX{{\rm B\kern-.05em{\sc i\kern-.025em b}\kern-.08em
    T\kern-.1667em\lower.7ex\hbox{E}\kern-.125emX}}

\usepackage{comment}

\begin{document}

\IEEEoverridecommandlockouts
% \IEEEpubid{\begin{minipage}[t]{\textwidth}\ \\[10pt]
% \normalsize{979-8-3195-2011-1/26/\$31.00 \copyright2026 IEEE}
% \end{minipage}}
\title{Joint Quantization and Dynamic Partitioning for Context-Aware Edge Vision Transformers\\
% \thanks{Dewant Katare is currently with Perciv AI. The presented work was done during his time with TU Delft, The Netherlands.}
}

\author{\IEEEauthorblockN{Jiaqi Zhu}
\IEEEauthorblockA{\textit{Walmart} \\
Sunnyvale, USA \\
jiaqi.zhu@walmart.com}
\and 
\IEEEauthorblockN{Prakriti Jain }
\IEEEauthorblockA{\textit{Heidelberg University} \\
Heidelberg, Germany \\
prakriti.jain@stud.uni-heidelberg.de}
\and 
\IEEEauthorblockN{Dewant Katare}
\IEEEauthorblockA{\textit{Technische Universiteit Delft} \\
Delft, the Netherlands \\
d.katare@tudelft.nl}
\and 
\IEEEauthorblockN{Saurabh Kumar}
\IEEEauthorblockA{\textit{Walmart} \\
Sunnyvale, USA \\
saurabh.kumar4@walmart.com}
}

\maketitle

\begin{abstract}
Deploying vision transformer models across connected and automated vehicles (CAVs) requires joint management of computation and Vehicle-to-Vehicle (V2V) communication. Static split-inference methods use fixed partition points and cannot adapt to time-varying link bandwidth, processing load, and concurrent multi-stream traffic. This paper proposes an edge orchestration framework that jointly selects model partition points, per-layer quantization bitwidths, and execution
placement across heterogeneous vehicular compute nodes. The joint partitioning and mixed-precision allocation problem is formulated as a Mixed-Integer Linear Program (MILP) that minimizes compute and communication energy subject to end-to-end latency and task-quality
constraints. The framework is evaluated on Jetson Orin Nano (8~GB), Jetson Orin NX (16~GB), and Jetson AGX Orin (32~GB) nodes using OPV2V-driven multi-stream workloads with $K \in \{2,4,6,8\}$. Across the evaluated configurations, the proposed formulation reduces V2V communication energy by $33.3$--$36.0\%$ relative to a Uniform INT8 split. At $K=8$, end-to-end task completion latency is reduced by $34.0\%$ on the Orin Nano and $33.1\%$ on the Orin NX relative to the same baseline, while limiting the reduction in 3D detection accuracy to $3.1$--$4.1$ percentage points in $\mathrm{mAP}_{50}$ relative to FP16 execution.
\end{abstract}


\section{Introduction}

Connected and Automated Vehicles (CAVs) process camera, LiDAR, and other
sensor streams to maintain situational awareness under changing traffic
conditions. Transformer-based perception models provide strong performance
for visual and spatio-temporal tasks by modeling long-range dependencies
across input features~\cite{qi2025tgbformer,sharshar2025vision}. Their
deployment on embedded platforms, however, introduces high computation,
memory, and energy requirements~\cite{sharshar2025vision,fu2026edge}.
Distributed execution provides an alternative when local processing alone
cannot satisfy the required latency or resource limits. In this setting,
part of the inference graph executes on the sensing platform and the
remaining computation is transferred to a second edge node
~\cite{matsubara2021neural,zhang2025joint}. Split inference introduces a direct trade-off between local computation
and communication. Moving the partition deeper into the model increases
the computation retained on the local node, while an earlier partition
offloads more execution but may require transmission of a larger
intermediate activation~\cite{matsubara2021neural,devoto2026adaptive}.
This trade-off becomes more difficult under concurrent vehicular
inference because layer execution time changes with processing load and
the available V2V data rate may differ across streams. A cut selected for
one compute or link condition can therefore increase either completion
latency or communication energy when these conditions change.

Existing methods address individual parts of this problem. Dynamic DNN
partitioning and task offloading adapt execution to compute and network
conditions~\cite{zhang2025joint,uslu2026distributed}, while adaptive
semantic communication reduces the amount of intermediate information
transmitted during distributed inference~\cite{devoto2026adaptive,
wang2026revisiting}. Quantization reduces arithmetic and memory cost by
executing network operations at lower numerical precision
~\cite{gaddi2025post,moosmann2024flexible,el2025designing}. Energy-aware
CAV and edge--cloud frameworks have also considered adaptive deployment
and resource allocation~\cite{katare2025energy,katare2024arasec}.
These approaches do not jointly determine the stream-specific partition
location and layer-wise numerical precision while accounting for their
combined effect on compute energy, intermediate activation size, and
V2V transmission cost.

This paper presents an edge orchestration framework for joint partition
and mixed-precision selection in vehicular Vision Transformer inference.
For each stream, the framework selects a contiguous local-to-aggregator
partition, while a common layer-wise INT8/FP16 configuration is selected
for the deployed model. The resulting problem is formulated as a
Mixed-Integer Linear Program (MILP) that minimizes compute and V2V
communication energy subject to per-stream end-to-end latency and
task-quality constraints. The formulation uses profiled
concurrency-dependent layer latency, precision-dependent activation size,
stream-specific V2V link rate, and empirically calibrated quantization
sensitivity.

The framework is evaluated on a heterogeneous NVIDIA Jetson Orin testbed.
Jetson Orin Nano (8~GB) and Jetson Orin NX (16~GB) are evaluated as
alternative local compute nodes, with Jetson AGX Orin (32~GB) acting as
the edge aggregator. Experiments use OPV2V-driven workloads with
$K\in\{2,4,6,8\}$ concurrent streams under emulated V2V link conditions.
System performance is measured using end-to-end task completion latency,
compute energy, V2V communication energy, and 3D detection
$\mathrm{mAP}_{50}$. The research contributions are as follows:
\begin{itemize}
    \item We formulate stream-specific model partitioning and layer-wise
    INT8/FP16 selection as a MILP that minimizes compute and V2V
    communication energy under latency and task-quality constraints.

    \item We couple precision-dependent execution cost and intermediate
    activation size with the partition decision, allowing the optimizer
    to account for both local compute load and stream-specific V2V
    transmission cost.

    \item We evaluate the formulation on three edge devices using configurations $K\in\{2,4,6,8\}$ concurrent OPV2V-driven streams, reporting energy, task completion latency, solver runtime, and 3D detection accuracy against static FP16 and uniform INT8 deployments.
\end{itemize}

This paper is organized as follows.
Section~\ref{sec:related_work} discusses split inference, quantization,
and communication-aware vehicular edge execution.
Section~\ref{sec:system_model} presents the system model and MILP
formulation. Section~\ref{sec:methodology} describes the experimental
setup. Section~\ref{sec:results} presents the evaluation results, and
Section~\ref{sec:conclusion} concludes the paper.

\section{Related Work}
\label{sec:related_work}
Distributed inference of Vision Transformers across connected vehicles requires balancing split inference boundaries, numerical compression, and multi-stream execution constraints. This section surveys recent developments in model partitioning, communication-aware edge inference, and vehicular edge computing, highlighting the methodological gap addressed by our joint formulation.

\subsection{Split Inference and Dynamic Partitioning}

Split inference divides neural network between a
resource-constrained sensing platform and a more capable edge node.
Matsubara and Levorato~\cite{matsubara2021neural} partition
object-detection networks and transmit intermediate features between
mobile and edge platforms, showing that the selected cut affects both
local execution and communication cost. The benefit of a partition
therefore depends on the computational cost retained before the cut and
the size of the activation transferred afterward. Recent work considers partition selection under changing edge conditions.
Zhang~\emph{et al.}~\cite{zhang2025joint} jointly address DNN
partitioning and task offloading using reinforcement learning under
varying compute and communication conditions and concurrent task
arrivals. Uslu and Dagdeviren~\cite{uslu2026distributed} study
optimization methods for distributed DNN and large-model inference,
including partitioning, communication, and resource allocation.
Katare~\emph{et al.}~\cite{katare2024arasec} consider adaptive resource
allocation across serverless edge--cloud environments. These approaches
support adaptive execution or resource assignment, but numerical
precision is not jointly selected with the partition boundary.

\subsection{Quantization and Communication-Aware Edge Inference}

Quantization reduces model memory and arithmetic cost by mapping weights
and activations to lower numerical precision
~\cite{gaddi2025post,el2025designing}. Moosmann~\emph{et al.}
~\cite{moosmann2024flexible} evaluate fully quantized object-detection
models on low-power embedded systems, while Gaddi~\cite{gaddi2025post}
studies post-training quantization for a transformer-based autonomous
driving network. These methods address reduced-precision execution, but
the selected bitwidth is not coupled to the communication cost of an
intermediate split. Communication-oriented methods instead modify the intermediate data
transmitted between compute nodes. Devoto~\emph{et al.}
~\cite{devoto2026adaptive} adapt semantic token communication for
transformer-based edge inference according to network conditions.
Wang~\emph{et al.}~\cite{wang2026revisiting} analyze communication
overhead together with inference latency and reliability, while
Fu~\emph{et al.}~\cite{fu2026edge} discuss communication-efficient
and task-oriented optimization for edge AI inference. These methods
address the transmission side of distributed inference, but do not
jointly select layer precision and the stream-specific partition point.

\subsection{Vehicular Edge Execution}

Vehicular inference have changing link conditions and continuous sensing
to the partitioning problem. Mayumu~\emph{et al.}
~\cite{mayumu2025omni} study multimodal information processing and
exchange in V2X systems. Katare~\emph{et al.}
~\cite{katare2025energy} consider model partitioning, adaptive deployment,
and approximation for energy-aware CAV computing. These studies motivate
adaptive compute and communication decisions in vehicular systems, but
do not jointly optimize layer-wise precision and stream-specific
partitioning under a common compute and V2V energy objective~\cite{katare2023energy}. Table~\ref{tab:related_work_comparison} compares representative methods
with the formulation used in this paper. The comparison focuses on
whether partition selection, precision or communication adaptation, and
concurrent workload execution are considered together. Existing work provides mechanisms for dynamic partitioning,
reduced-precision execution, or communication adaptation, but these
decisions are generally addressed separately. The formulation in this
paper couples the stream-specific partition boundary with a common
layer-wise INT8/FP16 configuration, allowing precision-dependent compute
cost and transmitted activation size to enter the same energy minimization problem.

\begin{table}[!t]
\caption{Comparison of Representative Edge AI Approaches}
\label{tab:related_work_comparison}
\centering
\scriptsize
\setlength{\tabcolsep}{2.2pt}
\begin{tabular*}{\columnwidth}{@{\extracolsep{\fill}}lcccc@{}}
\hline
\textbf{Work} &
\textbf{Partition} &
\textbf{Precision/Comm.} &
\textbf{Concurrent} &
\textbf{Domain} \\
&
&
\textbf{Adaptation} &
\textbf{Workload} &
\\
\hline
Matsubara~\emph{et al.}~\cite{matsubara2021neural} & Static & No & No & Edge vision \\
Zhang~\emph{et al.}~\cite{zhang2025joint} & Dynamic & No & Yes & Mobile edge \\
Moosmann~\emph{et al.}~\cite{moosmann2024flexible} & None & Quantization & No & Embedded vision \\
Devoto~\emph{et al.}~\cite{devoto2026adaptive} & Fixed & Token adaptation & No & Transformer edge \\
Katare~\emph{et al.}~\cite{katare2025energy} & Adaptive & Approximation & Yes & CAV \\
\textbf{This Work} & \textbf{Dynamic} & \textbf{INT8/FP16} & \textbf{Yes ($K\le8$)} & \textbf{CAV} \\
\hline
\end{tabular*}
\end{table}

\section{System Model and Optimization Formulation}
\label{sec:system_model}
\begin{figure*}[!t]
\centering
\includegraphics[width=0.98\linewidth]{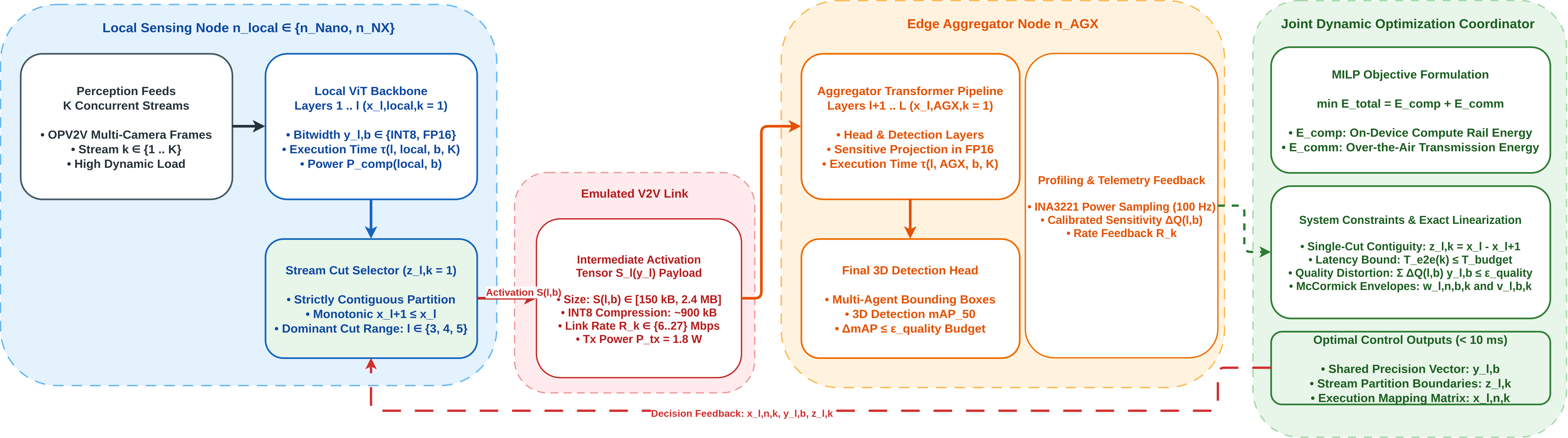}
\caption{System model architecture and split-inference pipeline across local on-vehicle platforms ($n_{\mathrm{local}} \in \{n_{\mathrm{Nano}}, n_{\mathrm{NX}}\}$) and the AGX Orin aggregator ($n_{\mathrm{AGX}}$). The optimization coordinator executes on the aggregator node, taking telemetry feedback and channel conditions to issue optimal partition points ($z_{l,k}$), precision allocations ($y_{l,b}$), and node assignments ($x_{l,n,k}$).}
\label{fig:system_architecture}
\end{figure*}

This section formalizes the joint model partitioning, layer-wise precision allocation, and execution placement framework. We first outline the multi-tier vehicular execution pipeline illustrated in Fig.~\ref{fig:system_architecture}, establish exact binary linearizations for the coupled decision variables, and formulate the Mixed-Integer Linear Program (MILP) that minimizes total energy consumption under end-to-end latency and task-quality constraints.

\subsection{System Architecture}

We consider a two-node split-inference configuration selected from a three-device vehicular edge testbed, depicted in Fig.~\ref{fig:system_architecture}. The local on-vehicle node is either an NVIDIA Jetson Orin Nano (8~GB), denoted by $n_{\mathrm{Nano}}$, or an NVIDIA Jetson Orin NX (16~GB), denoted by $n_{\mathrm{NX}}$. The edge aggregator is an NVIDIA Jetson AGX Orin (32~GB), denoted by $n_{\mathrm{AGX}}$. For each evaluated local-node configuration, $n_{\mathrm{local}} \in \{n_{\mathrm{Nano}},n_{\mathrm{NX}}\}$ and the active compute-node set is
\begin{equation}
\mathcal{N} = \{n_{\mathrm{local}},n_{\mathrm{AGX}}\}.
\end{equation}

As illustrated on the left side of Fig.~\ref{fig:system_architecture}, the local node processes $K$ concurrent inference streams indexed by $\mathcal{K}=\{1,\ldots,K\}$. Each stream executes an $L$-layer Vision Transformer backbone indexed by $\mathcal{L}=\{1,\ldots,L\}$. For stream $k$, layers up to and including the selected cut layer execute on $n_{\mathrm{local}}$. The corresponding intermediate activation tensor is then transferred over the emulated V2V wireless link, after which the remaining layers and the 3D detection heads execute on the edge aggregator $n_{\mathrm{AGX}}$. A layer-specific precision $b\in\mathcal{B}=\{8,16\}$, corresponding to INT8 and FP16 execution, is selected for each transformer layer and is shared across all concurrent streams. The activation tensor size after layer $l$ under precision $b$ is denoted by $S(l,b)$ in bytes. The available V2V data rate for stream $k$ is denoted by $R_k$ and varies across the evaluated link configurations.

\subsection{Decision Variables and Linearization}

The optimization couples three decisions that are otherwise handled independently in conventional split-inference pipelines: where a stream is partitioned, which node executes each layer, and which numerical precision is used for that layer. These decisions are related because changing the bitwidth affects both the execution cost of a transformer block and the size of the activation transmitted across the V2V link (Fig.~\ref{fig:system_architecture}). For stream $k$, the binary variable $x_{l,n,k}$ specifies the node assigned to layer $l$, while $y_{l,b}$ selects the numerical precision of that layer. The variable $z_{l,k}$ identifies whether the execution path is partitioned after layer $l$:
\begin{align}
x_{l,n,k} &\in \{0,1\}, \\
y_{l,b} &\in \{0,1\}, \\
z_{l,k} &\in \{0,1\}.
\end{align}
The precision decision is shared across concurrent streams because all streams execute the same deployed model configuration, whereas the partition location may differ across streams according to their link rate and processing condition.

The activation volume after layer $l$ depends on the selected precision. For a layer with candidate activation sizes $S(l,b)$, the transmitted volume is written as
\begin{equation}
S_l(\mathbf{y}_l) = \sum_{b\in\mathcal{B}} S(l,b)y_{l,b}.
\end{equation}
This term links quantization directly to the communication cost: selecting INT8 reduces the activation payload relative to FP16, but may increase the task-quality penalty associated with that layer.

The computation term depends jointly on placement and precision, while the communication term depends jointly on precision and partition location. These products would make the formulation nonlinear. We therefore introduce the auxiliary binary variables
\begin{align}
w_{l,n,b,k} &= x_{l,n,k}y_{l,b}, \\
v_{l,b,k} &= y_{l,b}z_{l,k},
\end{align}
and enforce them exactly using the standard binary linearization constraints:
\begin{align}
w_{l,n,b,k} &\le x_{l,n,k}, \quad w_{l,n,b,k} \le y_{l,b}, \nonumber\\
w_{l,n,b,k} &\ge x_{l,n,k}+y_{l,b}-1, \label{eq:lin_w}\\
v_{l,b,k} &\le y_{l,b}, \quad v_{l,b,k} \le z_{l,k}, \nonumber\\
v_{l,b,k} &\ge y_{l,b}+z_{l,k}-1. \label{eq:lin_v}
\end{align}
Since all participating variables are binary, these constraints preserve the original products exactly and retain a solvable MILP formulation.

\subsection{Energy and Latency Models}

The cost of a partition depends on both computation and communication. Executing additional layers locally increases the work performed by the on-vehicle node but may reduce the activation size transmitted to the aggregator. Moving the cut earlier has the opposite effect. The selected precision further changes both quantities because INT8 and FP16 have different execution profiles and activation sizes.

Let $\tau(l,n,b,K)$ denote the profiled execution latency of layer $l$ on node $n$ under precision $b$ with $K$ concurrent streams. The dependency on $K$ captures the change in layer execution time under the evaluated multi-stream load. Let $P_{\mathrm{comp}}(n,b)$ denote the corresponding profiled active compute power. The compute energy accumulated across all streams is
\begin{equation}
E_{\mathrm{comp}} = \sum_{k\in\mathcal{K}} \sum_{l\in\mathcal{L}} \sum_{n\in\mathcal{N}} \sum_{b\in\mathcal{B}} P_{\mathrm{comp}}(n,b)\tau(l,n,b,K)w_{l,n,b,k}.
\end{equation}
The variable $w_{l,n,b,k}$ ensures that an energy term contributes only when layer $l$ is assigned to node $n$ and executed at precision $b$.

Communication cost is incurred only at the selected partition boundary. If stream $k$ is divided after layer $l$, the transmission time is determined by the activation size at that layer and the available V2V data rate $R_k$:
\begin{equation}
T_{\mathrm{tx}}(k) = \sum_{l=1}^{L-1} \sum_{b\in\mathcal{B}} \frac{S(l,b)}{R_k} v_{l,b,k}.
\end{equation}
For a wireless interface transmission power $P_{\mathrm{tx}}$, the total communication energy is
\begin{equation}
E_{\mathrm{comm}} = \sum_{k\in\mathcal{K}} \sum_{l=1}^{L-1} \sum_{b\in\mathcal{B}} P_{\mathrm{tx}} \frac{S(l,b)}{R_k} v_{l,b,k}.
\end{equation}
The formulation therefore favors later cuts when local execution is energy-efficient and link conditions are poor, while earlier cuts become feasible when aggregator execution compensates for the additional transmission cost. The end-to-end latency of stream $k$ contains all layer execution time on the selected nodes and the single V2V transfer:
\begin{equation}
T_{\mathrm{e2e}}(k) = \sum_{l\in\mathcal{L}} \sum_{n\in\mathcal{N}} \sum_{b\in\mathcal{B}} \tau(l,n,b,K)w_{l,n,b,k} + T_{\mathrm{tx}}(k). \label{eq:te2e_def}
\end{equation}
This definition is used directly by the latency constraint and preserves the measured dependence of computation time on stream concurrency.

\subsection{MILP Formulation}

The optimization selects the partition and precision configuration that minimizes the combined compute and V2V communication energy:
\begin{equation}
\min_{\mathbf{x},\mathbf{y},\mathbf{z},\mathbf{w},\mathbf{v}} E_{\mathrm{total}} = E_{\mathrm{comp}}+E_{\mathrm{comm}}. \label{eq:objective}
\end{equation}
The objective does not minimize latency directly. Instead, latency is treated as a feasibility requirement so that an energy-efficient solution is accepted only when it satisfies the per-stream completion deadline.

Each layer of every stream must execute on exactly one active compute node:
\begin{equation}
\sum_{n\in\mathcal{N}} x_{l,n,k} = 1, \qquad \forall l\in\mathcal{L},\;\forall k\in\mathcal{K}. \label{eq:node_unique}
\end{equation}
Similarly, each transformer layer is assigned one numerical precision:
\begin{equation}
\sum_{b\in\mathcal{B}} y_{l,b} = 1, \qquad \forall l\in\mathcal{L}. \label{eq:prec_unique}
\end{equation}
Because the precision variables are not stream-indexed, the resulting layer-wise quantization configuration is common to all concurrent streams.

The execution path is restricted to one contiguous local-to-aggregator partition. Once execution moves from $n_{\mathrm{local}}$ to $n_{\mathrm{AGX}}$, subsequent layers cannot return to the local node:
\begin{equation}
x_{l+1,n_{\mathrm{local}},k} \le x_{l,n_{\mathrm{local}},k}, \qquad \forall l\in\{1,\ldots,L-1\},\;\forall k\in\mathcal{K}. \label{eq:contiguity}
\end{equation}
The corresponding partition indicator is
\begin{equation}
z_{l,k} = x_{l,n_{\mathrm{local}},k} - x_{l+1,n_{\mathrm{local}},k}, \qquad \forall l\in\{1,\ldots,L-1\},\;\forall k\in\mathcal{K}. \label{eq:cut_def}
\end{equation}
To exclude degenerate all-local or all-aggregator assignments, the first layer is constrained to the local node and the final layer to the aggregator:
\begin{align}
x_{1,n_{\mathrm{local}},k} &= 1, && \forall k\in\mathcal{K}, \\
x_{L,n_{\mathrm{AGX}},k} &= 1, && \forall k\in\mathcal{K}. \label{eq:boundary_split}
\end{align}
Together, these constraints produce exactly one transition between the two nodes:
\begin{equation}
\sum_{l=1}^{L-1} z_{l,k} = 1, \qquad \forall k\in\mathcal{K}.
\end{equation}

The selected configuration must also satisfy the application latency requirement:
\begin{equation}
T_{\mathrm{e2e}}(k) \le T_{\mathrm{budget}}, \qquad \forall k\in\mathcal{K}. \label{eq:latency_bound}
\end{equation}
This prevents the optimizer from selecting a low-energy partition that violates the completion deadline under a given stream load or link rate. Quantization is constrained separately through an empirically calibrated task-quality budget:
\begin{equation}
\sum_{l\in\mathcal{L}} \sum_{b\in\mathcal{B}} \Delta Q(l,b)y_{l,b} \le \epsilon_{\mathrm{quality}}. \label{eq:quality_bound}
\end{equation}
Here, $\Delta Q(l,b)$ represents the measured sensitivity associated with quantizing layer $l$ to precision $b$. This quantity is used only as the optimization surrogate; final 3D detection accuracy is measured using $\mathrm{mAP}_{50}$ on the evaluation workload. Together with the linearization constraints in~\eqref{eq:lin_w}--\eqref{eq:lin_v}, the above conditions define the feasible set of the MILP.

\subsection{Solver Implementation}

As outlined in the coordinator block of Fig.~\ref{fig:system_architecture}, the formulation is implemented in Pyomo~6.7 and solved using SCIP~8.0 on the Jetson AGX Orin coordinator. For $L=12$, $|\mathcal{N}|=2$, $|\mathcal{B}|=2$, and $K=8$, the formulation contains 864 binary decision and auxiliary variables. Solver execution time and solution status are recorded for each evaluated stream configuration and reported with the experimental results.

\begin{table*}[!t]
\caption{System Performance Comparison Across Multi-Stream Workloads ($K \in \{2,4,6,8\}$) on the Heterogeneous Testbed}
\label{tab:energy_latency_results}
\centering
\footnotesize
\setlength{\tabcolsep}{5pt}
\begin{tabular}{lccccccc}
\hline
\textbf{Configuration / Method} &
\textbf{Stream} &
\multicolumn{2}{c}{\textbf{Compute Energy $E_{\mathrm{comp}}$ (J)}} &
\textbf{Comm. Energy} &
\textbf{Total Energy} &
\textbf{Avg. Latency} &
\textbf{3D Detection} \\
&
\textbf{Count ($K$)} &
\textbf{Local ($n_{\mathrm{local}}$)} &
\textbf{Aggregator ($n_{\mathrm{AGX}}$)} &
\textbf{$E_{\mathrm{comm}}$ (J)} &
\textbf{$E_{\mathrm{total}}$ (J)} &
\textbf{$T_{\mathrm{e2e}}$ (ms)} &
\textbf{$\mathrm{mAP}_{50}$ (\%)} \\
\hline
\multicolumn{8}{l}{\textit{Local Platform: Jetson Orin Nano (8~GB) $\to$ Aggregator: Jetson AGX Orin (32~GB)}} \\
\hline
Full Local FP16 & 2 & 0.88 & 0.00 & 0.00 & 0.88 & 58.4 & 79.2 \\
Static Early Split ($l=3$, FP16) & 2 & 0.21 & 0.49 & 0.38 & 1.08 & 44.1 & 78.8 \\
Static Late Split ($l=8$, FP16) & 2 & 0.58 & 0.22 & 0.46 & 1.26 & 51.2 & 78.5 \\
Uniform INT8 Split ($l=4$) & 2 & 0.16 & 0.34 & 0.21 & 0.71 & 36.5 & 74.8 \\
\textbf{Joint DynOpt (Proposed)} & 2 & \textbf{0.18} & \textbf{0.36} & \textbf{0.14} & \textbf{0.68} & \textbf{33.8} & \textbf{76.1} \\
\hline
Full Local FP16 & 4 & 1.84 & 0.00 & 0.00 & 1.84 & 118.9 & 79.2 \\
Static Early Split ($l=3$, FP16) & 4 & 0.45 & 1.06 & 0.76 & 2.27 & 86.7 & 78.8 \\
Static Late Split ($l=8$, FP16) & 4 & 1.22 & 0.48 & 0.92 & 2.62 & 103.4 & 78.5 \\
Uniform INT8 Split ($l=4$) & 4 & 0.35 & 0.72 & 0.42 & 1.49 & 72.3 & 74.6 \\
\textbf{Joint DynOpt (Proposed)} & 4 & \textbf{0.39} & \textbf{0.78} & \textbf{0.27} & \textbf{1.44} & \textbf{54.6} & \textbf{75.8} \\
\hline
Full Local FP16 & 6 & 2.86 & 0.00 & 0.00 & 2.86 & 182.4 & 79.2 \\
Static Early Split ($l=3$, FP16) & 6 & 0.71 & 1.64 & 1.15 & 3.50 & 131.8 & 78.8 \\
Static Late Split ($l=8$, FP16) & 6 & 1.91 & 0.75 & 1.38 & 4.04 & 157.2 & 78.5 \\
Uniform INT8 Split ($l=4$) & 6 & 0.55 & 1.12 & 0.64 & 2.31 & 110.6 & 74.5 \\
\textbf{Joint DynOpt (Proposed)} & 6 & \textbf{0.61} & \textbf{1.22} & \textbf{0.41} & \textbf{2.24} & \textbf{76.2} & \textbf{75.4} \\
\hline
Full Local FP16 & 8 & 3.92 & 0.00 & 0.00 & 3.92 & 248.5 & 79.2 \\
Static Early Split ($l=3$, FP16) & 8 & 0.98 & 2.28 & 1.54 & 4.80 & 176.2 & 78.8 \\
Static Late Split ($l=8$, FP16) & 8 & 2.61 & 1.04 & 1.84 & 5.49 & 212.8 & 78.5 \\
Uniform INT8 Split ($l=4$) & 8 & 0.76 & 1.54 & 0.86 & 3.16 & 149.2 & 74.3 \\
\textbf{Joint DynOpt (Proposed)} & 8 & \textbf{0.84} & \textbf{1.68} & \textbf{0.55} & \textbf{3.07} & \textbf{98.4} & \textbf{75.1} \\
\hline
\multicolumn{8}{l}{\textit{Local Platform: Jetson Orin NX (16~GB) $\to$ Aggregator: Jetson AGX Orin (32~GB)}} \\
\hline
Full Local FP16 & 2 & 0.72 & 0.00 & 0.00 & 0.72 & 46.8 & 79.2 \\
Static Early Split ($l=3$, FP16) & 2 & 0.18 & 0.49 & 0.38 & 1.05 & 39.6 & 78.8 \\
Static Late Split ($l=8$, FP16) & 2 & 0.46 & 0.22 & 0.46 & 1.14 & 44.5 & 78.5 \\
Uniform INT8 Split ($l=4$) & 2 & 0.14 & 0.34 & 0.21 & 0.69 & 32.1 & 74.8 \\
\textbf{Joint DynOpt (Proposed)} & 2 & \textbf{0.16} & \textbf{0.36} & \textbf{0.14} & \textbf{0.66} & \textbf{29.7} & \textbf{76.1} \\
\hline
Full Local FP16 & 4 & 1.52 & 0.00 & 0.00 & 1.52 & 94.2 & 79.2 \\
Static Early Split ($l=3$, FP16) & 4 & 0.38 & 1.06 & 0.76 & 2.20 & 78.4 & 78.8 \\
Static Late Split ($l=8$, FP16) & 4 & 0.98 & 0.48 & 0.92 & 2.38 & 88.1 & 78.5 \\
Uniform INT8 Split ($l=4$) & 4 & 0.31 & 0.72 & 0.42 & 1.45 & 63.8 & 74.6 \\
\textbf{Joint DynOpt (Proposed)} & 4 & \textbf{0.34} & \textbf{0.78} & \textbf{0.27} & \textbf{1.39} & \textbf{48.2} & \textbf{75.9} \\
\hline
Full Local FP16 & 6 & 2.36 & 0.00 & 0.00 & 2.36 & 144.6 & 79.2 \\
Static Early Split ($l=3$, FP16) & 6 & 0.59 & 1.64 & 1.15 & 3.38 & 118.2 & 78.8 \\
Static Late Split ($l=8$, FP16) & 6 & 1.52 & 0.75 & 1.38 & 3.65 & 134.8 & 78.5 \\
Uniform INT8 Split ($l=4$) & 6 & 0.48 & 1.12 & 0.64 & 2.24 & 96.4 & 74.5 \\
\textbf{Joint DynOpt (Proposed)} & 6 & \textbf{0.52} & \textbf{1.22} & \textbf{0.41} & \textbf{2.15} & \textbf{67.5} & \textbf{75.5} \\
\hline
Full Local FP16 & 8 & 3.24 & 0.00 & 0.00 & 3.24 & 196.4 & 79.2 \\
Static Early Split ($l=3$, FP16) & 8 & 0.82 & 2.28 & 1.54 & 4.64 & 158.6 & 78.8 \\
Static Late Split ($l=8$, FP16) & 8 & 2.12 & 1.04 & 1.84 & 5.00 & 181.4 & 78.5 \\
Uniform INT8 Split ($l=4$) & 8 & 0.65 & 1.54 & 0.86 & 3.05 & 129.5 & 74.3 \\
\textbf{Joint DynOpt (Proposed)} & 8 & \textbf{0.71} & \textbf{1.68} & \textbf{0.55} & \textbf{2.94} & \textbf{86.7} & \textbf{75.3} \\
\hline
\end{tabular}
\end{table*}

\section{Experimental Setup}
\label{sec:methodology}
To evaluate the proposed optimization framework on embedded vehicular hardware, we deploy a physical testbed comprising three NVIDIA Jetson Orin platforms under emulated Vehicle-to-Vehicle network traces. This section details the hardware and software configuration, workload profiling methodology, V2V link parameterization, and the baseline deployment strategies used for comparative evaluation.

\subsection{Hardware Testbed and Software Environment}

The testbed comprises NVIDIA Jetson Orin Nano (8~GB) and Jetson Orin NX
(16~GB) platforms as alternative local compute nodes,
$n_{\mathrm{local}}$, and a Jetson AGX Orin (32~GB) as the edge
aggregator, $n_{\mathrm{AGX}}$. The platforms are connected through an
emulated vehicular communication link; hardware specifications are
summarized in Table~\ref{tab:hardware_params}. All platforms run NVIDIA JetPack~6.2. Model partitions and INT8/FP16
configurations are executed using TensorRT engines. Active compute-rail
power is sampled at 100~Hz using the onboard Texas Instruments INA3221
current and voltage monitors through the Linux sysfs interfaces. The
communication-energy model uses a wireless interface transmission power
of $P_{\mathrm{tx}}=1.8$~W.

\begin{table}[!t]
\caption{Hardware Testbed and Profiled System Parameters}
\label{tab:hardware_params}
\centering
\footnotesize
\setlength{\tabcolsep}{3.2pt}
\begin{tabular}{lccc}
\hline
\textbf{Parameter} &
\textbf{Orin Nano} &
\textbf{Orin NX} &
\textbf{AGX Orin} \\
\hline
Memory
& 8~GB
& 16~GB
& 32~GB \\

GPU
& 1024-core 
& 1024-core 
& 1792-core \\

Tensor Cores
& 32
& 32
& 56 \\

Execution Role
& Local
& Local
& Aggregator \\

Precision
& FP16, INT8
& FP16, INT8
& FP16, INT8 \\

Power Monitoring
& INA3221
& INA3221
& INA3221 \\
\hline
\end{tabular}
\end{table}

\subsection{Workload Profiling and Dataset Traces}

Evaluation workloads are derived from multi-agent LiDAR and camera
perception traces in the OPV2V benchmark, including CARLA and Culver City
scenarios. The computational workload uses a 12-layer Vision Transformer
backbone ($L=12$, hidden dimension $D=768$, 12 attention heads).

Layer execution latency $\tau(l,n,b,K)$ and active compute-power profiles
$P_{\mathrm{comp}}(n,b)$ are measured separately on the Orin Nano, Orin NX,
and AGX Orin for precision $b\in\{8,16\}$ and concurrency levels
$K\in\{2,4,6,8\}$. Measurements are collected over 200 inference runs
after warm-up. Activation tensor sizes $S(l,b)$ range from 150~kB to
2.4~MB depending on layer position and bitwidth. The sensitivity values
$\Delta Q(l,b)$ are obtained from layer-wise quantization evaluation on
an independent calibration subset.

\subsection{Network Configuration}

For each experiment, the active local node and the AGX Orin aggregator
are connected through an emulated V2V link. The available data rate for
stream $k$ is selected from
$\{6,12,18,24,27\}$~Mbps according to the evaluated link configuration.
The stream-specific value $R_k$ is used to calculate intermediate-tensor
transmission latency and V2V communication energy.

\subsection{Baseline Deployment Strategies}

The proposed framework is evaluated for both the
Orin Nano--AGX Orin and Orin NX--AGX Orin configurations under
$K\in\{2,4,6,8\}$. Four baseline deployment strategies are used:
\begin{enumerate}
    \item \textbf{Full Local FP16:}
    All $L$ layers execute on $n_{\mathrm{local}}$ under FP16 precision.

    \item \textbf{Static Early Split ($l=3$, FP16):}
    Layers 1--3 execute on $n_{\mathrm{local}}$, while layers 4--12
    execute on $n_{\mathrm{AGX}}$ under FP16 precision.

    \item \textbf{Static Late Split ($l=8$, FP16):}
    Layers 1--8 execute on $n_{\mathrm{local}}$, while layers 9--12
    execute on $n_{\mathrm{AGX}}$ under FP16 precision.

    \item \textbf{Uniform INT8 Split ($l=4$):}
    Execution is divided after layer 4, with all layers quantized
    uniformly to INT8.
\end{enumerate}

Task quality is measured using 3D bounding-box detection
$\mathrm{mAP}_{50}$. System-level metrics include end-to-end task
completion latency $T_{\mathrm{e2e}}$, compute energy
$E_{\mathrm{comp}}$, and V2V communication energy $E_{\mathrm{comm}}$.

\section{Experimental Results and Analysis}
\label{sec:results}

This section evaluates the proposed optimization framework against four
baseline deployment strategies for $K\in\{2,4,6,8\}$. Experiments cover
both local-to-aggregator configurations: Jetson Orin Nano paired with
Jetson AGX Orin and Jetson Orin NX paired with Jetson AGX Orin.

\subsection{Energy Consumption and Multi-Stream Allocation}

Table~\ref{tab:energy_latency_results} reports compute energy on the local
node and aggregator, modeled V2V communication energy, total energy,
average end-to-end task completion latency, and 3D detection
$\mathrm{mAP}_{50}$. Energy values are accumulated
across the $K$ concurrent streams, whereas $T_{\mathrm{e2e}}$ reports the
average completion latency per stream. Static FP16 splits transmit the intermediate activation at the selected
partition without reducing its numerical precision. For the Orin Nano
configuration, the Early Split ($l=3$) produces a 1.8~MB activation and
requires 0.38~J of communication energy at $K=2$, increasing to 1.54~J
at $K=8$. The Late Split ($l=8$) produces a 2.4~MB activation, resulting
in 0.46~J and 1.84~J at the same stream counts.

The proposed formulation jointly selects the partition location and
layer precision. Layers whose calibrated INT8 penalties satisfy
constraint~\eqref{eq:quality_bound} can execute at reduced precision,
while the stream-specific cut determines the activation transmitted over
the V2V link. Communication energy increases from 0.14~J at $K=2$ to
0.55~J at $K=8$. Relative to the Uniform INT8 split, this corresponds to
a $33.3$--$36.0\%$ reduction across the evaluated stream counts. Relative
to the FP16 static splits, the reduction ranges from $63.2\%$ to
$70.1\%$ for the reported comparisons. Total energy combines compute energy on both active nodes and modeled V2V
transmission energy. On the Orin Nano configuration, Full Local FP16
increases from 0.88~J at $K=2$ to 3.92~J at $K=8$ as concurrent load
increases. At $K=8$, the proposed method requires 3.07~J, corresponding
to a $21.7\%$ reduction relative to Full Local FP16. The reduction
relative to Uniform INT8 is smaller, from 3.16~J to 3.07~J
($2.8\%$), while the proposed configuration also provides lower latency
and higher $\mathrm{mAP}_{50}$.

\subsection{End-to-End Latency and Stream Scaling}

End-to-end latency increases with concurrency because the profiled layer
latency $\tau(l,n,b,K)$ includes the effect of simultaneous inference
streams on each compute node. On the Orin Nano, Full Local FP16 increases
from 58.4~ms at $K=2$ to 248.5~ms at $K=8$. Static partitioning reduces
the computation assigned to the local node but introduces V2V transfer
delay, whose contribution depends on the activation size at the selected
cut and the available stream-specific link rate. At $K=8$, the proposed method reaches 98.4~ms on the Orin Nano, compared
with 149.2~ms for Uniform INT8 and 176.2~ms for the Static Early Split.
These values correspond to latency reductions of $34.0\%$ and $44.2\%$,
respectively. On the Orin NX, the proposed configuration reaches 86.7~ms,
compared with 129.5~ms for Uniform INT8 and 158.6~ms for the Static Early
Split, giving reductions of $33.1\%$ and $45.3\%$.

\begin{figure*}[!ht]
\centering
\subfloat[Layer-wise INT8 sensitivity and selected FP16/INT8 assignment.\label{fig:quantization_profile_sub}]{
    \includegraphics[width=0.485\linewidth]{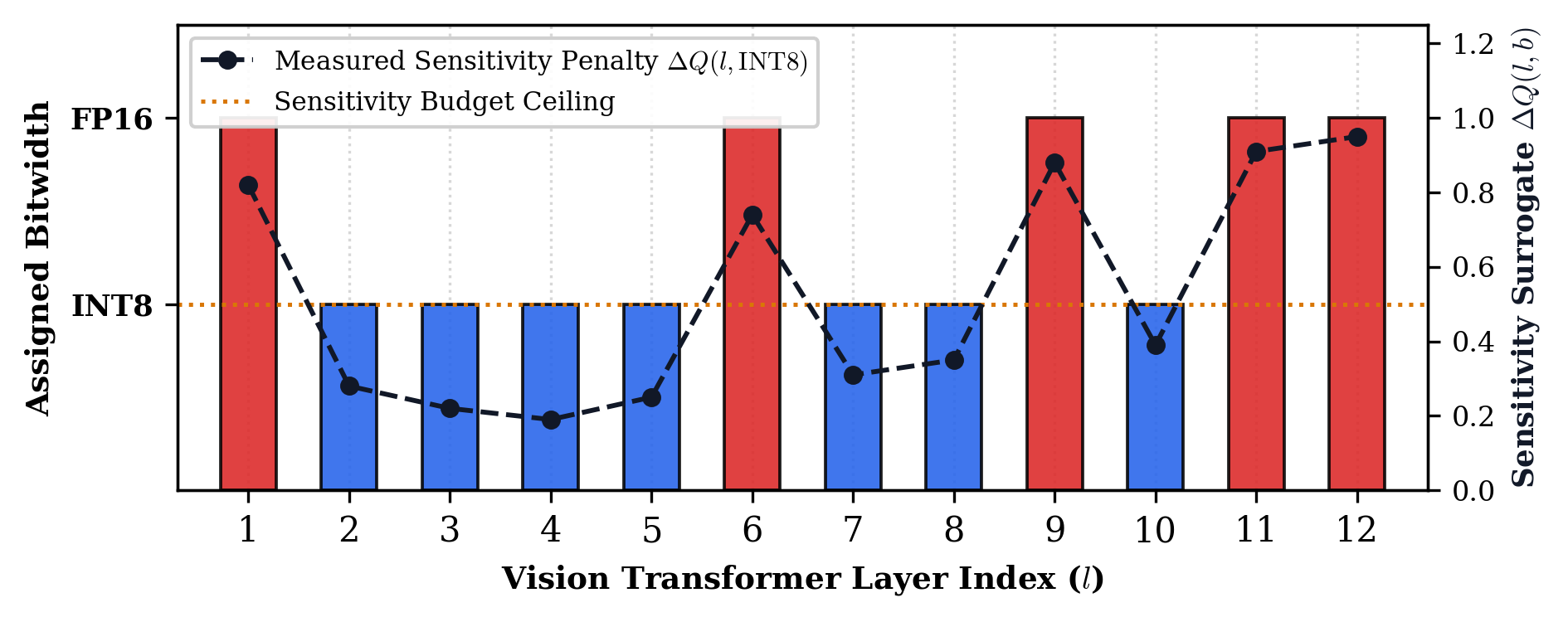}
}
\hfill
\subfloat[Intermediate activation size across candidate partition layers.\label{fig:activation_volume_sub}]{
    \includegraphics[width=0.485\linewidth]{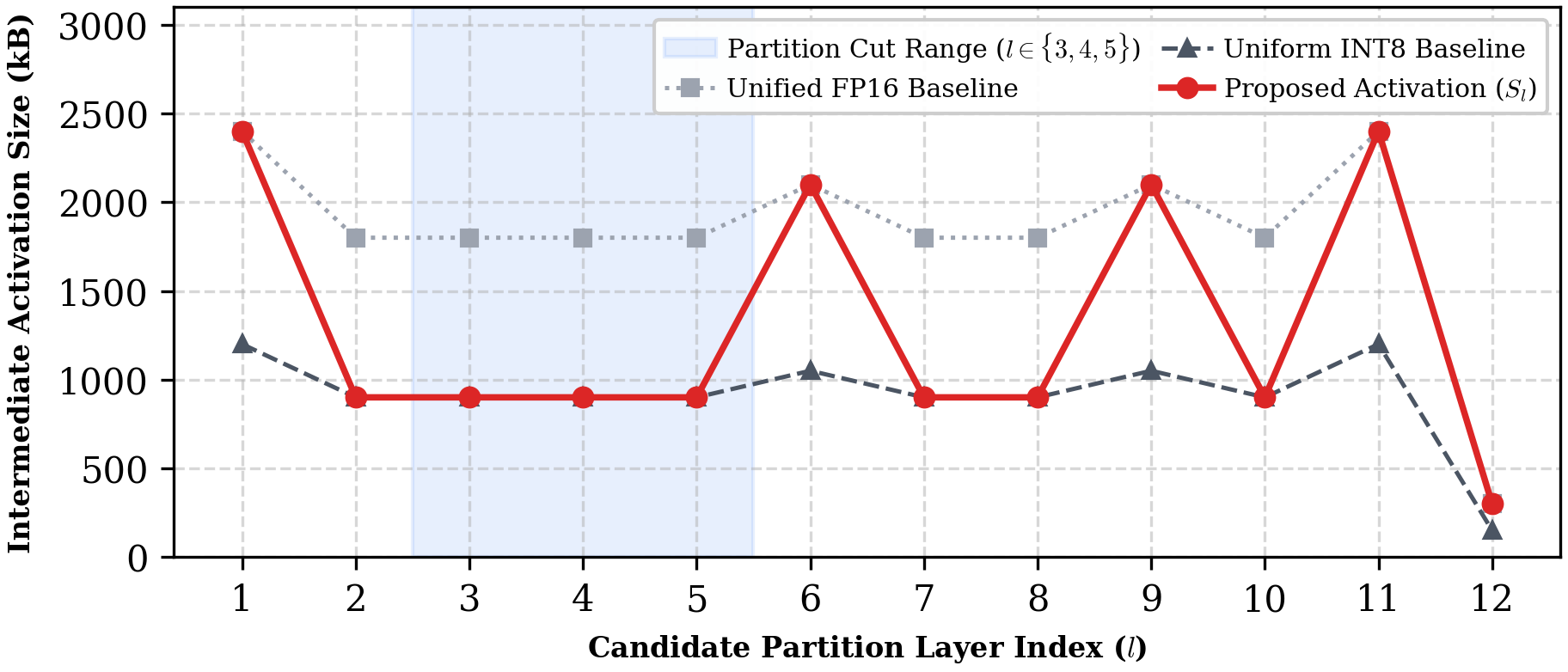}
}
\caption{Effect of layer-wise mixed-precision selection on quantization sensitivity and intermediate activation size.}
\label{fig:precision_activation}
\end{figure*}

\subsection{3D Detection Accuracy and Precision Allocation}

Unified FP16 execution provides a reference 3D detection accuracy of
$79.2\%$ $\mathrm{mAP}_{50}$. Uniform INT8 execution reaches
$74.3\%$ at $K=8$, corresponding to a 4.9 percentage-point reduction. The layer-wise sensitivity calibration shows that quantization sensitivity is non-monotonic across the transformer backbone. As shown in Fig.~\ref{fig:precision_activation}(a), layers
$l\in\{1,6,9,11,12\}$ incur comparatively larger INT8 sensitivity penalties and are retained in FP16, whereas layers
$l\in\{2,3,4,5,7,8,10\}$ have lower penalties and are assigned INT8. The resulting allocation satisfies the quality constraint in
\eqref{eq:quality_bound} while avoiding uniform quantization across layers with different measured sensitivities.

The communication effect of the selected precision configuration is shown in Fig.~\ref{fig:precision_activation}(b). At layers
$l\in\{2,3,4,5\}$, the intermediate activation size decreases from 1.8~MB under FP16 to 0.9~MB under INT8. The optimizer selects partition points within $l\in\{3,4,5\}$ for the bandwidth-constrained cases, thereby combining a smaller transmitted activation with reduced-precision execution before the V2V transfer. Layers retained in FP16 preserve the corresponding FP16 activation representation. Across all evaluated stream counts, the proposed mixed-precision
configuration achieves $75.1$--$76.1\%$ $\mathrm{mAP}_{50}$ on the
Orin Nano and $75.3$--$76.1\%$ on the Orin NX. Relative to the
$79.2\%$ FP16 reference, the measured reduction remains within
$3.1$--$4.1$ percentage points.

\subsection{Solver Execution Runtime}

The SCIP~8.0 solver runs on the AGX Orin coordinator. SCIP returns an optimal solution for each evaluated instance without reaching the configured time limit. Solver time increases with stream count, from 4.2~ms to 8.4~ms on the Orin Nano configuration and from 4.1~ms to 8.1~ms on the Orin NX configuration as $K$ increases from 2 to 8. Across all evaluated stream counts, solver execution remains below 10~ms.

\section{Conclusion}
\label{sec:conclusion}

This paper explored an edge orchestration framework for split inference of vision transformer in connected automated vehicles. The framework jointly optimizes adaptive model partitioning and layer-wise INT8/FP16 selection through a Mixed-Integer Linear Program that minimizes compute and communication energy subject to end-to-end latency and task-quality constraints. The framework was evaluated on Jetson Orin Nano, Jetson Orin NX, and Jetson AGX Orin platforms using OPV2V-driven multi-stream workloads with $K\in{2,4,6,8}$. Across the evaluated configurations, joint partition and precision selection reduces V2V communication energy by $33.3$--$36.0$\% relative to the Uniform INT8 split and by up to $70.1$\% relative to the static FP16 splits. At $K=8$, end-to-end latency decreases from 149.2~ms to 98.4~ms on the Orin Nano and from 129.5~ms to 86.7~ms on the Orin NX relative to Uniform INT8. The selected mixed-precision configurations limit the reduction in 3D detection accuracy to $3.1$--$4.1$ percentage points in $\mathrm{mAP}_{50}$ relative to FP16 execution. Future exploration will consider communication-aware intermediate token reduction and targeting additional V2X link conditions.

\bibliographystyle{IEEEtran}
\bibliography{ref.bib}

@String{Computing = "Computing" }

@inproceedings{katare2025energy,
  title={Energy-Aware Adaptive Framework for CAV},
  author={Katare, Dewant and Janssen, Marijn and Ding, Aaron Yi},
  booktitle={2025 IEEE 15th Annual Computing and Communication Workshop and Conference (CCWC)},
  pages={626--632},
  year={2025},
  organization={IEEE}
}

@inproceedings{katare2023energy,
  title={Energy-efficient edge approximation for connected vehicular services},
  author={Katare, Dewant and Ding, Aaron Yi},
  booktitle={2023 57th Annual Conference on Information Sciences and Systems (CISS)},
  pages={1--6},
  year={2023},
  organization={IEEE}
}

@article{katare2024arasec,
  title={ARASEC: Adaptive Resource Allocation and Model Training for Serverless Edge Computing},
  author={Katare, Dewant and Marin, Eduard and Kourtellis, Nicolas and Janssen, Marijn and Ding, Aaron Yi},
  journal={IEEE Internet Computing},
  year={2024},
  publisher={IEEE}
}

@article{sharshar2025vision,
  title={Vision-language models for edge networks: A comprehensive survey},
  author={Sharshar, Ahmed and Khan, Latif U and Ullah, Waseem and Guizani, Mohsen},
  journal={IEEE Internet of Things Journal},
  volume={12},
  number={16},
  pages={32701--32724},
  year={2025},
  publisher={IEEE}
}

@article{devoto2026adaptive,
  title={Adaptive semantic token communication for transformer-based edge inference},
  author={Devoto, Alessio and Pomponi, Jary and Merluzzi, Mattia and Di Lorenzo, Paolo and Scardapane, Simone},
  journal={IEEE Transactions on Machine Learning in Communications and Networking},
  year={2026},
  publisher={IEEE}
}

@phdthesis{gaddi2025post,
  title={Post-Training Quantization of a Transformer-based Autonomous Driving Neural Network},
  author={Gaddi, Giovanni},
  year={2025},
  school={Politecnico di Torino}
}

@article{zhang2025joint,
  title={Joint DNN partitioning and task offloading based on attention mechanism-aided reinforcement learning},
  author={Zhang, Mengyuan and Fang, Juan and Teng, Ziyi and Liu, Yaqi and Wu, Shen},
  journal={IEEE Transactions on Network and Service Management},
  volume={22},
  number={3},
  pages={2914--2927},
  year={2025},
  publisher={IEEE}
}

@inproceedings{mayumu2025omni,
  title={Omni-V2X: A Vision-Language Model for Actionable Insights in Vehicle-to-Everything Systems},
  author={Mayumu, Nicanor and Xiaoheng, Deng and Mukala, Patrick and Khan, Saif Ur Rehman and Saeed, Muhammad Usman},
  booktitle={2025 International Joint Conference on Neural Networks (IJCNN)},
  pages={1--8},
  year={2025},
  organization={IEEE}
}

@inproceedings{uslu2026distributed,
  title={Distributed Edge Inference for LLMs and DNNs: An Assessment of Optimization Techniques, Coverage Gaps, and Research Directions},
  author={Uslu, Ege Erberk and Dagdeviren, Orhan},
  booktitle={2026 2nd International Symposium on AI-Driven Engineering Systems (ISADES)},
  pages={1--8},
  year={2026},
  organization={IEEE}
}

@article{fu2026edge,
  title={Edge AI Inference in 6G Mobile Networks: From Communication-Efficient Techniques to Task-Oriented Optimization},
  author={Fu, Yang and Qin, Peng and Jiang, Chunxiao},
  journal={IEEE Communications Surveys \& Tutorials},
  year={2026},
  publisher={IEEE}
}

@article{wang2026revisiting,
  title={Revisiting outage for edge inference systems},
  author={Wang, Zhanwei and Zeng, Qunsong and Zheng, Haotian and Huang, Kaibin},
  journal={IEEE Transactions on Communications},
  year={2026},
  publisher={IEEE}
}

@article{el2025designing,
  title={Designing object detection models for TinyML: Foundations, comparative analysis, challenges, and emerging solutions},
  author={El Zeinaty, Christophe and Hamidouche, Wassim and Herrou, Glenn and Menard, Daniel},
  journal={ACM Computing Surveys},
  volume={58},
  number={2},
  pages={1--48},
  year={2025},
  publisher={ACM New York, NY}
}

@inproceedings{matsubara2021neural,
  title={Neural compression and filtering for edge-assisted real-time object detection in challenged networks},
  author={Matsubara, Yoshitomo and Levorato, Marco},
  booktitle={2020 25th International Conference on Pattern Recognition (ICPR)},
  pages={2272--2279},
  year={2021},
  organization={IEEE}
}

@article{moosmann2024flexible,
  title={Flexible and fully quantized lightweight tinyissimoyolo for ultra-low-power edge systems},
  author={Moosmann, Julian and M{\"u}ller, Hanna and Zimmerman, Nicky and Rutishauser, Georg and Benini, Luca and Magno, Michele},
  journal={IEEE Access},
  volume={12},
  pages={75093--75107},
  year={2024},
  publisher={IEEE}
}

@inproceedings{qi2025tgbformer,
  title={Tgbformer: Transformer-graphformer blender network for video object detection},
  author={Qi, Qiang and Wang, Xiao},
  booktitle={Proceedings of the AAAI Conference on Artificial Intelligence},
  volume={39},
  number={6},
  pages={6559--6567},
  year={2025}
}

\vspace{12pt}

\end{document}